\documentclass[10pt,twocolumn,letterpaper]{article}

\usepackage[pagenumbers]{cvpr} % To force page numbers, e.g. for an arXiv version
\usepackage{caption}
\usepackage{multirow}
\usepackage{graphicx}

\makeatletter
\g@addto@macro\@maketitle{%
    \begin{center}
        \includegraphics[width=\textwidth]{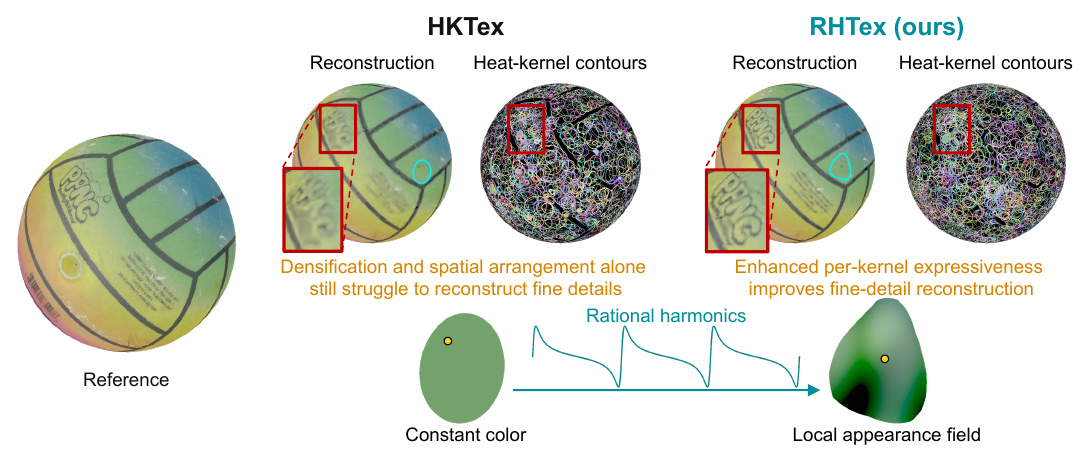}
        \captionsetup{hypcap=false}
        \captionof{figure}{\textbf{RHTex (Rational Harmonic Textures)} equips each heat kernel with a spatially varying local appearance field represented by rational trigonometric atoms, enabling more faithful reconstruction of high-frequency surface details.}
        \label{fig:teaser}
    \end{center}
}
\makeatother

\definecolor{cvprblue}{rgb}{0.21,0.49,0.74}
\usepackage[pagebackref,breaklinks,colorlinks,allcolors=cvprblue]{hyperref}

\def\confName{CVPR}
\def\confYear{2027}

\title{Rational Harmonic Textures: High-Frequency Surface Appearance with Heat Kernels}

\author{%
  Jing Li\textsuperscript{1}\quad
  Yawei Luo\textsuperscript{2,*}\quad
  Xiangze Meng\textsuperscript{1}\quad
  Ying Li\textsuperscript{3}\quad
  Jinwu Gao\textsuperscript{1}\\
  Yi Chang\textsuperscript{1}\quad
  Huajun Chen\textsuperscript{2}\quad
  Tieru Wu\textsuperscript{1}\quad
  Rui Ma\textsuperscript{1,*}\\[3pt]
  \textsuperscript{1}\,Jilin University\quad
  \textsuperscript{2}\,Zhejiang University\quad
  \textsuperscript{3}\,North China University of Technology\\
  \textsuperscript{*}\,Corresponding authors
}

\begin{document}
\maketitle
\begin{abstract}
Heat kernel textures provide an intrinsic representation of surface appearance grounded in discrete Riemannian geometry. However, existing formulations assign a constant color to each kernel, limiting per-kernel expressiveness. Kernel densification and spatial arrangement alone may not fully compensate for this limitation when reconstructing high-frequency details. To address this, we introduce Rational Harmonic Textures, which equip each heat kernel with a compact, spatially varying local appearance field represented by learnable rational trigonometric atoms. Each atom represents a harmonic series with geometrically decaying amplitudes, enriching within-kernel color variation while retaining intrinsic heat-kernel coverage on the surface. During adaptive densification, kernel splitting creates child kernels with new centers and potentially different local tangent frames. Directly copying a parent’s appearance parameters to its children can misalign the inherited appearance fields and introduce abrupt changes during optimization. We therefore further derive an analytic reparameterization that jointly adjusts wave vectors and phase offsets to compensate for center shifts and tangent-frame changes when initializing split children. Experiments demonstrate that our method achieves better surface texture reconstruction than the existing heat kernel texture representation.
\end{abstract}    
\section{Introduction}
\label{sec:intro}

Accurately representing surface appearance is essential for reconstructing and rendering detailed 3D assets. Fine patterns, thin markings, and rapid color variations require texture representations that capture high-frequency signals while respecting the geometry of the underlying surface. Conventional texture maps store this information in a two-dimensional atlas, requiring a surface parameterization that can introduce seams, distortion, and uneven sampling density. These limitations have motivated alternatives that avoid a global UV atlas by representing appearance as continuous fields or associating color and features directly with the surface~\cite{yuksel2010mesh,oechsle2019texture,koestler2022intrinsic,mahajan2024meshfeat}. For UV-free representations, a key challenge is to allocate local representational capacity adaptively while respecting surface geometry.

More recently, Heat Kernel Textures (HKTex)~\cite{foti2026heat} addresses this challenge with an explicit, UV-free texture representation based on anisotropic heat kernels, which act as geodesic counterparts of Euclidean Gaussians and naturally conform to the underlying surface. The representation is coupled with Riemannian optimization and manifold-aware density control, enabling kernel motion and adaptive refinement directly on the mesh. Despite these advantages, each heat kernel models only the spatial diffusion of a single color over the surface. Texture variation therefore depends on the spatially varying contributions of overlapping kernels. Fitting a texture requires coordinating the number and spatial distribution of the kernels. Although adaptive densification can increase local kernel density, it does not increase the expressive capacity of individual kernels. Consequently, reconstructing complex high-frequency details remains challenging under a limited kernel budget, as illustrated in \cref{fig:teaser}.

Motivated by this observation, we introduce \textbf{Rational Harmonic Textures}, which equip each heat kernel with a compact, spatially varying local appearance field. The field is expressed using four rational trigonometric atoms evaluated in the kernel's local tangent frame. Each atom encodes a harmonic series with geometrically decaying coefficient magnitudes, parameterized by a learnable frequency, phase offset, color coefficients, and a shape parameter that controls the harmonic weights. This formulation allows a single kernel to capture richer appearance variations across the surface while retaining heat-kernel-based spatial coverage. The resulting representation provides greater capacity for modeling high-frequency details under the same kernel budget.

Building on this, we further propose \emph{frame-aware harmonic reparameterization} to initialize the local appearance fields of child kernels during adaptive densification. Kernel splitting creates child kernels with new centers and potentially different local tangent frames. Since each local appearance field is defined relative to its kernel center and tangent frame, directly copying the parent's appearance parameters can misalign the inherited fields. Such misalignment can perturb the learned local appearance, leaving subsequent optimization to correct the resulting discrepancies. We therefore derive analytic updates that project parent wave vectors onto the child tangent planes and adjust phase offsets to compensate for center translations. The resulting initialization locally aligns each child's appearance field with the parent's under a tangent-plane approximation. Our key contributions are as follows:

\begin{itemize}
    \item We introduce Rational Harmonic Textures, a surface appearance representation that equips each heat kernel with a compact local appearance field.
    
    \item We derive an analytic reparameterization that jointly adjusts wave vectors and phase offsets during splitting, compensating for center translations and tangent-frame changes to provide locally consistent appearance.
    
    \item Experiments demonstrate that our method captures high-frequency surface details more effectively than the original heat kernel texture representation.
\end{itemize}

%-------------------------------------------------------------------------

\section{Related work}
\label{sec:RW}

%-------------------------------------------------------------------------

\paragraph{Surface appearance representations.}
Surface appearance can be stored explicitly on mesh elements or represented by continuous functions. Per-face textures~\cite{burley2008ptex} and mesh-associated color samples~\cite{yuksel2010mesh} avoid a global UV atlas, while neural representations~\cite{oechsle2019texture} use coordinate-based color fields, learned texture mappings~\cite{xiang2021neutex}, or mesh-attached features~\cite{yang2022neumesh,mahajan2024meshfeat}. Surface-aligned sparse voxel fields~\cite{wang2026beyond} have also been explored for texture compression. Intrinsic formulations~\cite{koestler2022intrinsic,elhag2024manifold} use Laplace--Beltrami eigenfunctions to encode surface points for neural texture reconstruction and generative modeling of manifold signals. Related texture synthesis and generation methods~\cite{kovacs2024surface,mitchel2024single,foti2024uv} exploit surface convolutions, tangent-vector latent representations, or heat-based surface operators. HKTex~\cite{foti2026heat} represents appearance using anisotropic surface heat kernels, with Riemannian optimization and adaptive density control. Subsequent work~\cite{he2026accelerating} explores local unfolding and randomized thermal features as alternatives to global spectral evaluation, reducing preprocessing and fitting costs. Our method builds on HKTex by replacing each kernel's constant color with a compact rational harmonic appearance field, while retaining its heat-kernel evaluation and spatial coverage.

\begin{figure*}[t]
    \centering
    \includegraphics[width=\linewidth]{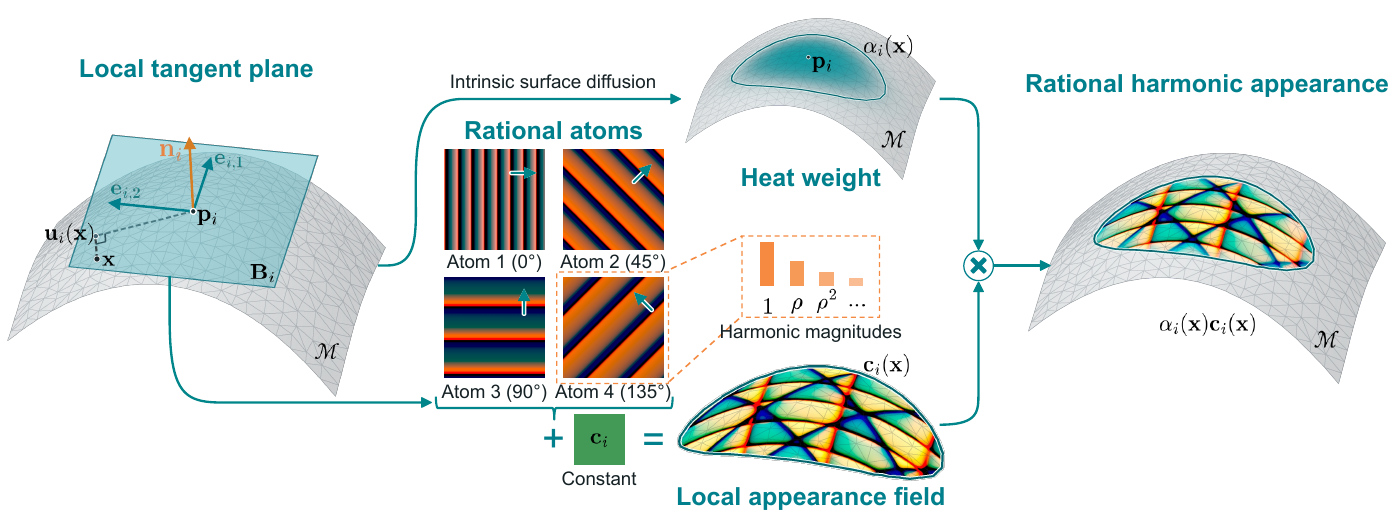}
    \caption{\textbf{Rational harmonic appearance.} We map surface points $\mathbf{x}$ to local tangent coordinates $\mathbf{u}_i(\mathbf{x})$ defined at the kernel center $\mathbf{p}_i$. A learnable per-kernel color offset $\mathbf{c}_i$ combines with four rational trigonometric atoms to form the local appearance field $\mathbf{c}_i(\mathbf{x})$. Each atom represents a harmonic series with geometrically decaying amplitudes. Arrows indicate the four initial wave-vector directions. Multiplying the local appearance field by the heat weight $\alpha_i(\mathbf{x})$ yields the single-kernel contribution $\alpha_i(\mathbf{x})\mathbf{c}_i(\mathbf{x})$.}
    \label{fig:pipeline}
\end{figure*}

\paragraph{Local appearance modeling.}
To capture fine detail without relying solely on primitive densification, existing methods introduce spatially varying appearance through per-primitive texture maps~\cite{rong2025gstex,chao2025textured,papantonakis2025content,svitov2025billboard}, local basis functions or small neural networks~\cite{huang2025textured,xu2026svgs}, and shared neural representations for local textures or surface-adjacent appearance fields~\cite{wang2025neural,zhang2025neural}. Frequency-based formulations further enrich local color functions~\cite{watanabe20253d,Watanabe_2026_CVPR}, encode interpolated features harmonically~\cite{condor2026neural}, or modulate spatial compositing kernels~\cite{10.1145/3746027.3754855}. During optimization, child appearance can be initialized through parent-texture sampling or neural-parameter inheritance~\cite{papantonakis2025content,Watanabe_2026_CVPR}. These approaches primarily enhance local appearance in Gaussian splatting representations. Extending such models from Euclidean primitives to surface heat kernels requires accommodating the geometry of curved surfaces. We formulate rational harmonic fields using local tangent-plane approximations and analytically transfer their wave vectors and phase offsets during kernel splitting to account for center shifts and tangent-frame changes.

%-------------------------------------------------------------------------

\section{Preliminaries}
\label{sec:pre}

\paragraph{Heat kernel textures.}
HKTex~\cite{foti2026heat} represents surface appearance using a collection of $N$ anisotropic heat kernels defined on a triangular mesh $\mathcal{M}$. The $i$-th kernel has a center $\mathbf{p}_i\in\mathcal{M}$, parameters controlling its spatial distribution, and a learnable color $\mathbf{c}_i$, where $i=1,\ldots,N$. The heat diffusion associated with each kernel is computed using the spectral properties of an anisotropic Laplace--Beltrami operator~\cite{andreux2014anisotropic}. For a surface point $\mathbf{x}\in\mathcal{M}$, the heat kernel centered at $\mathbf{p}_i$ can be approximated by a truncated spectral expansion:
\begin{equation}
    h_i(\mathbf{x})
    \approx
    \sum_{k=1}^{K}
    e^{-t\lambda_{i,k}}\,
    \varphi_{i,k}(\mathbf{x})\,
    \varphi_{i,k}(\mathbf{p}_i)\,
    m_i,
    \label{eq:heat_kernel}
\end{equation}
where $t$ is a fixed diffusion time, $K$ is the number of retained spectral components, and $\lambda_{i,k}$ and $\varphi_{i,k}$ denote the $k$-th eigenvalue and eigenfunction associated with the $i$-th kernel's anisotropic diffusion parameters. The scalar $m_i$ denotes the mass at the kernel center. The spectral quantities associated with each kernel are interpolated from precomputed eigensystems according to its anisotropy and orientation parameters. Eigenfunction values at kernel centers and query points are further evaluated by barycentric interpolation within the containing mesh triangles.

The heat response is normalized by its value at the kernel center and weighted according to the biharmonic distance to suppress contributions from distant regions. The weighted response is then passed through a rescaled soft-step function with a learnable threshold and sharpness parameter for each kernel, producing the weight $\alpha_i(\mathbf{x})$. For each surface point, candidate kernels are selected based on biharmonic distance. Let $\mathcal{I}(\mathbf{x})\subseteq\{1,\ldots,N\}$ denote the indices of the $K_c$ candidates with the largest filtered weights. Their colors are combined according to these weights and added to a learnable global base color $\mathbf{c}_{\mathrm{base}}$:
\begin{equation}
    \mathbf{C}(\mathbf{x})
    =
    \mathbf{c}_{\mathrm{base}}
    +
    \frac{
        \sum_{i\in\mathcal{I}(\mathbf{x})}
        \alpha_i(\mathbf{x})\,\mathbf{c}_i
    }{
        \max\!\left(
            1,\,
            \sum_{i\in\mathcal{I}(\mathbf{x})}
            \alpha_i(\mathbf{x})
        \right)
    }.
    \label{eq:hktex_color}
\end{equation}

Optimization constrains kernels to the mesh surface. We retain HKTex's heat-kernel weighting and adaptive density-control settings while enriching per-kernel appearance.

%-------------------------------------------------------------------------

\section{Methodology}
\label{sec:Method}
We first introduce our rational harmonic appearance representation (\cref{fig:pipeline}) in \cref{sec:rha}. We then derive a frame-aware harmonic reparameterization for initializing child appearance fields during kernel splitting in \cref{sec:fcat}. Finally, we describe the optimization objective in \cref{sec:optimization}.

\subsection{Rational harmonic appearance}
\label{sec:rha}

\begin{figure*}[t]
    \centering
    \includegraphics[width=\linewidth]{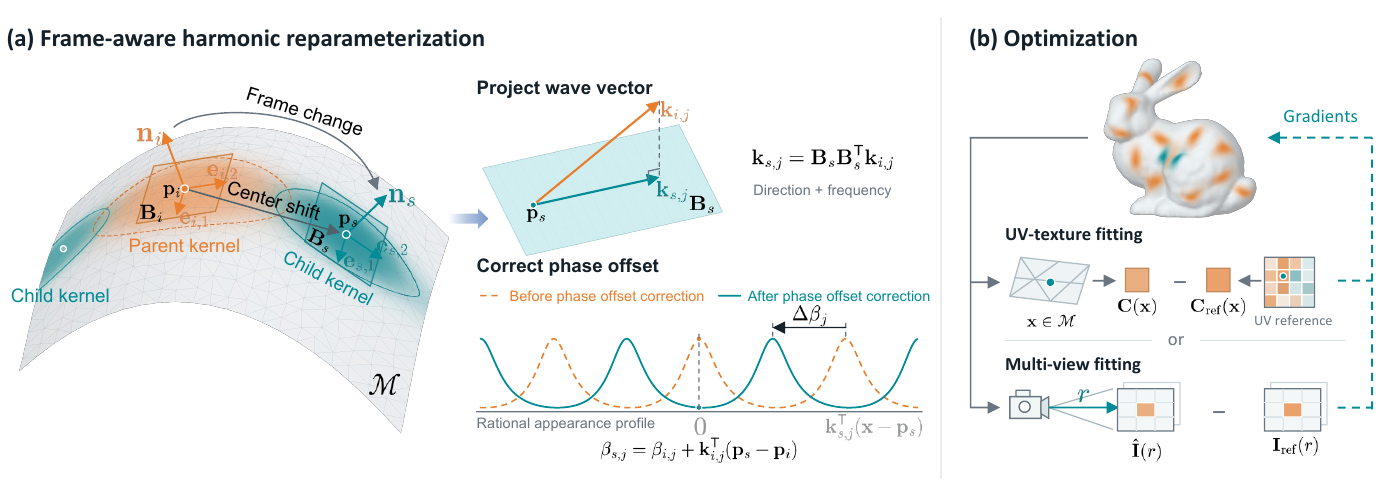}
    \caption{Overview of frame-aware harmonic reparameterization and optimization. (a) During kernel splitting, parent wave vectors are projected onto each child's tangent plane to update wave directions and frequencies, while phase offsets are corrected for displacements from the parent center. These updates align the parent and child harmonic fields near each child center under a tangent-plane approximation. (b) Kernel parameters are optimized using either UV-texture fitting or multi-view fitting.}
    \label{fig:frame_transfer}
\end{figure*}

Given the intrinsic, geodesic nature of heat kernels, local geodesic coordinates provide a natural starting point for modeling spatially varying surface appearance. Such coordinates can be obtained through the Riemannian logarithmic map $\log_{\mathbf{p}_i}(\mathbf{x})$, which represents a nearby surface point as a tangent vector encoding the geodesic distance and the initial direction of the shortest geodesic from $\mathbf{p}_i$ to $\mathbf{x}$~\cite{sharp2019vector}. However, repeatedly evaluating these maps as kernel centers move during optimization would incur additional computational cost. We therefore adopt tangent-plane projection as a local approximation~\cite{schmidt2006interactive}, expressing the displacement from each kernel center in its local tangent frame.

For the $i$-th kernel, we define a local tangent-plane approximation at $\mathbf{p}_i$ using a unit normal obtained by normalizing the barycentric interpolation of area-weighted vertex normals. Let $\mathbf{B}_i = [\mathbf{e}_{i,1}\;\mathbf{e}_{i,2}] \in\mathbb{R}^{3\times 2}$ contain two orthonormal basis vectors spanning this plane. Projecting the displacement from the kernel center onto this basis and normalizing by a fixed length scale $\ell_i>0$ gives the local coordinates
\begin{equation}
    \mathbf{u}_i(\mathbf{x})
    =
    \frac{1}{\ell_i}
    \mathbf{B}_i^{\mathsf T}
    (\mathbf{x}-\mathbf{p}_i),
    \label{eq:local_coordinates}
\end{equation}
where $\ell_i$ sets the length unit of the appearance coordinates. All kernels are initialized with $\ell_i=\sqrt{A_{\mathcal{M}}/(\pi N_0)}$, where $A_{\mathcal{M}}$ is the total mesh surface area and $N_0$ is the initial number of kernels. This scale remains fixed during optimization and is inherited by newly created kernels. As the kernel center moves, the tangent basis is updated using the minimal rotation that aligns the previous and current interpolated normals.

Using these local coordinates, we represent each kernel's appearance with four rational trigonometric atoms, indexed by $j=1,\ldots,4$. The phase of the $j$-th atom is
\begin{equation}
    \psi_{i,j}(\mathbf{x})
    =
    2\pi f_{i,j}\,
    \mathbf{d}_{i,j}^{\mathsf T}\mathbf{u}_i(\mathbf{x})
    + \beta_{i,j},
    \label{eq:wave_phase}
\end{equation}
where $\mathbf{d}_{i,j}\in\mathbb{R}^2$ is a unit direction, $f_{i,j}>0$ is a learnable frequency in normalized local coordinates, and $\beta_{i,j}\in\mathbb{R}$ is a learnable phase offset. The directions are initialized at the four prescribed angles $(j-1)\pi/4$ and are excluded from gradient-based optimization. They are updated analytically when initializing split children, as described in \cref{sec:fcat}.

Each atom comprises a rational cosine component and a rational sine component:
\begin{equation}
\begin{aligned}
    F(\psi,\rho)
    &=
    \frac{\cos\psi-\rho}
         {1-2\rho\cos\psi+\rho^2},\\
    G(\psi,\rho)
    &=
    \frac{\sin\psi}
         {1-2\rho\cos\psi+\rho^2},
\end{aligned}
\label{eq:rational_components}
\end{equation}
where $\rho$ controls the relative weights of the harmonics. For $|\rho|<1$, these rational functions admit the Fourier series expansions
\begin{equation}
\begin{aligned}
    F(\psi,\rho)
    &= \sum_{n=1}^{\infty}\rho^{n-1}\cos(n\psi),\\
    G(\psi,\rho)
    &= \sum_{n=1}^{\infty}\rho^{n-1}\sin(n\psi).
\end{aligned}
\label{eq:rational_harmonics}
\end{equation}
The $n$-th harmonic has weight $\rho^{n-1}$, whose magnitude $|\rho|^{n-1}$ decays geometrically with harmonic order. When $\rho=0$, only the fundamental harmonic remains, yielding $F(\psi,0)=\cos\psi$ and $G(\psi,0)=\sin\psi$. Nonzero $\rho$ introduces higher-order harmonics, enabling richer local variation without independently parameterizing each harmonic. The closed-form expressions evaluate these infinite series without explicit harmonic summation. For each atom, we parameterize $\rho_{i,j}=0.8\tanh(r_{i,j})$, where $r_{i,j}\in\mathbb{R}$ is learnable. This ensures $|\rho_{i,j}|<0.8$ and keeps the denominator in \cref{eq:rational_components} uniformly bounded away from zero.

The local appearance field combines a learnable per-kernel color offset with spatially varying rational harmonic components:
\begin{equation}
\begin{aligned}
    \mathbf{c}_i(\mathbf{x})
    &= \mathbf{c}_i
    + \sum_{j=1}^{4}
      \mathbf{a}_{i,j}
      F\big(\psi_{i,j}(\mathbf{x}),\rho_{i,j}\big)\\
    &\quad
    + \sum_{j=1}^{4}
      \mathbf{b}_{i,j}
      G\big(\psi_{i,j}(\mathbf{x}),\rho_{i,j}\big),
\end{aligned}
\label{eq:local_appearance}
\end{equation}
where $\mathbf{c}_i\in\mathbb{R}^3$ is the learnable per-kernel color offset introduced in \cref{sec:pre}, and $\mathbf{a}_{i,j},\mathbf{b}_{i,j}\in\mathbb{R}^3$ are learnable RGB coefficient vectors for the rational cosine and sine components, respectively.

The surface color is obtained by substituting $\mathbf{c}_i(\mathbf{x})$ for $\mathbf{c}_i$ in \cref{eq:hktex_color}. The heat-kernel weights $\alpha_i(\mathbf{x})$ determine the spatial contribution of each local field, while the rational atoms model color variation within that contribution.

\subsection{Frame-aware harmonic reparameterization}
\label{sec:fcat}

Introducing spatial variation within each kernel also requires reconsidering how appearance is inherited during adaptive densification. In the original HKTex representation, each kernel's color parameter is independent of local coordinates and can therefore be copied directly to its children. In contrast, our rational harmonic appearance field depends on both the kernel center and its local tangent frame. Since splitting changes the center and may alter the tangent frame, directly copying the parent's parameters can misalign the inherited appearance field. Such misalignment can perturb the learned local appearance, requiring subsequent optimization to correct these discrepancies. We therefore derive an analytic reparameterization of wave vectors and phase offsets that jointly accounts for center translations and tangent-frame changes, as illustrated in \cref{fig:frame_transfer}.

\begin{figure}[t]
    \centering
    \includegraphics[width=\linewidth]{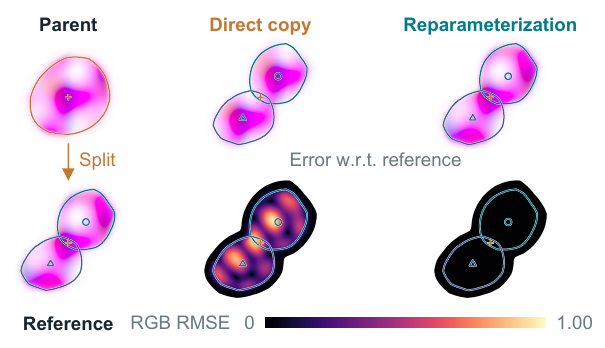}
    \caption{Frame-aware harmonic reparameterization better matches the reference than direct parameter copying. The reference uses the parent appearance field; all three share child heat weights and blend normalization. Error maps show pointwise RGB RMSE relative to the reference. The cross marks the parent center; the circle and triangle mark child centers.}
    \label{fig:split}
\end{figure}

Consider a parent kernel $i$ and one of its children $s$, with center $\mathbf{p}_s$, tangent basis $\mathbf{B}_s$, and inherited coordinate scale $\ell_s=\ell_i$. The child tangent basis is first updated using the interpolated surface normal at its new center. To account for differences between the parent and child frames, we first express the parent's phase variation in a common three-dimensional coordinate system. Substituting \cref{eq:local_coordinates} into \cref{eq:wave_phase} yields
\begin{equation}
\begin{aligned}
    \mathbf{k}_{i,j}
    &= \frac{2\pi f_{i,j}}{\ell_i}
       \mathbf{B}_i\mathbf{d}_{i,j},\\
    \psi_{i,j}(\mathbf{x})
    &= \mathbf{k}_{i,j}^{\mathsf T}
       (\mathbf{x}-\mathbf{p}_i)+\beta_{i,j}.
\end{aligned}
\label{eq:ambient_wave_vector}
\end{equation}
Here, $\mathbf{k}_{i,j}\in\mathbb{R}^3$ is the wave vector of the parent's $j$-th atom, representing its spatial phase gradient in ambient three-dimensional space.

\begin{table*}[t]
    \centering
    \small
    \setlength{\tabcolsep}{4pt}
    \caption{Quantitative results for UV-texture fitting and multi-view supervision. HKTex and RHTex use the same maximum kernel count: 5,000 for UV fitting and 10,000 for multi-view fitting. Best results are bold; second-best results are underlined.}
    \label{tab:uv_multiview_results}
    \begin{tabular}{lccccccc}
        \toprule
        Method
        & \shortstack{PSNR \\ $(\uparrow)$}
        & \shortstack{LPIPS \\ $(\times 10^{-2}, \downarrow)$}
        & \shortstack{SSIM \\ $(\times 10^{-2}, \uparrow)$}
        & \shortstack{MS-SSIM \\ $(\times 10^{-2}, \uparrow)$}
        & \shortstack{MSE \\ $(\times 10^{-3}, \downarrow)$}
        & \shortstack{Time
          \\ $(\mathrm{min}, \downarrow)$}
        & \shortstack{Storage
          \\ $(\mathrm{KB}, \downarrow)$} \\
        \midrule
        \multicolumn{8}{l}{\textit{UV texture fitting}} \\

        MLP Pos. Enc.~\cite{foti2026heat}
        & 39.460 & 2.106 & 98.051 & 98.427 & 0.586
        & 5.61 & 487.695 \\

        InstantNGP~\cite{muller2022instant}
        & 40.225 & 2.071 & 98.115 & 98.591 & \underline{0.441}
        & 4.82 & \underline{482.964} \\

        INFs~\cite{koestler2022intrinsic}
        & 39.380 & 2.298 & 97.933 & 98.431 & 0.585
        & 4.63 & 494.143 \\

        HKTex~\cite{foti2026heat}
        & \underline{40.297} & \underline{1.863}
        & \underline{98.208} & \underline{98.715} & 0.447
        & \textbf{2.57} & \textbf{97.334} \\
        
        \midrule
        RHTex (Ours)
        & \textbf{40.748} & \textbf{1.302} & \textbf{98.595}
        & \textbf{98.854} & \textbf{0.420}
        & \underline{2.87} & 522.411 \\

        \midrule
        \multicolumn{8}{l}{\textit{Multi-view reconstruction}} \\

        MLP Pos. Enc.~\cite{foti2026heat}
        & 33.510 & 5.160 & 95.846 & 95.200 & 2.263
        & 117.85 & \underline{381.033} \\

        NVDiffRec$^{*}$~\cite{munkberg2022extracting}
        & \underline{35.553} & 3.581 & \underline{96.976}
        & 97.081 & 0.998
        & 114.81 & 384.902 \\

        HKTex~\cite{foti2026heat}
        & 35.232 & \underline{3.445} & 96.939
        & \underline{97.607} & \underline{0.689}
        & \textbf{92.29} & \textbf{84.445} \\
        
        \midrule
        RHTex (Ours)
        & \textbf{35.646} & \textbf{2.598} & \textbf{97.624}
        & \textbf{97.960} & \textbf{0.619}
        & \underline{96.21} & 413.436 \\

        \bottomrule
    \end{tabular}
\end{table*}

\begin{table*}[t]
    \centering
    \caption{UV-texture fitting under different kernel limits. Avg.\ kernels denotes the mean final kernel count. Best results are bold; second-best results are underlined.}
    \label{tab:uv_kernel_limit}
    \begin{tabular}{lccccccc}
        \toprule
        Method
        & \shortstack{Max Kernels\\limit}
        & \shortstack{Avg.\\kernels}
        & \shortstack{PSNR\\$(\uparrow)$}
        & \shortstack{LPIPS\\$(\times 10^{-2}, \downarrow)$}
        & \shortstack{SSIM\\$(\times 10^{-2}, \uparrow)$}
        & \shortstack{MS-SSIM\\$(\times 10^{-2}, \uparrow)$}
        & \shortstack{MSE\\$(\times 10^{-3}, \downarrow)$} \\
        \midrule
        \multirow{6}{*}{HKTex~\cite{foti2026heat}}
        & 5000 & 4834
        & 40.297 & 1.863 & 98.208 & 98.715 & 0.447 \\
        & 6000 & 5701
        & \underline{40.386} & 1.826 & 98.238
        & \underline{98.739} & \underline{0.442} \\
        & 7000 & 6521
        & 40.329 & 1.817 & 98.245 & 98.732 & 0.445 \\
        & 8000 & 7251
        & 40.355 & 1.804 & 98.255 & 98.738 & 0.445 \\
        & 9000 & 7943
        & 40.347 & \underline{1.791} & 98.259 & 98.738 & 0.445 \\
        & 10000 & 8578
        & 40.349 & 1.798 & \underline{98.261}
        & \underline{98.739} & 0.445 \\
        \midrule
        RHTex (Ours)
        & 5000 & 4898
        & \textbf{40.748} & \textbf{1.302} & \textbf{98.595}
        & \textbf{98.854} & \textbf{0.420} \\
        \bottomrule
    \end{tabular}
\end{table*}

% \begin{table*}[t]
%     \centering
%     \caption{UV-texture fitting under different kernel limits. Avg.~kernels denotes the mean final kernel count. Best reconstruction results are highlighted in bold.}
%     \label{tab:uv_kernel_limit}
%     \begin{tabular}{lccccccc}
%         \toprule
%         Method
%         & \shortstack{Max Kernels\\limit ($\downarrow$)}
%         & \shortstack{Avg.\\kernels ($\downarrow$)}
%         & \shortstack{PSNR\\$(\uparrow)$}
%         & \shortstack{LPIPS\\$(\times 10^{-2}, \downarrow)$}
%         & \shortstack{SSIM\\$(\times 10^{-2}, \uparrow)$}
%         & \shortstack{MS-SSIM\\$(\times 10^{-2}, \uparrow)$}
%         & \shortstack{MSE\\$(\times 10^{-3}, \downarrow)$} \\
%         \midrule
%         \multirow{6}{*}{HKTex~\cite{foti2026heat}}
%         & 5000 & 4834
%         & 40.297 & 1.863 & 98.208 & 98.715 & 0.447 \\
%         & 6000 & 5701
%         & 40.386 & 1.826 & 98.238 & 98.739 & 0.442 \\
%         & 7000 & 6521
%         & 40.329 & 1.817 & 98.245 & 98.732 & 0.445 \\
%         & 8000 & 7251
%         & 40.355 & 1.804 & 98.255 & 98.738 & 0.445 \\
%         & 9000 & 7943
%         & 40.347 & 1.791 & 98.259 & 98.738 & 0.445 \\
%         & 10000 & 8578
%         & 40.349 & 1.798 & 98.261 & 98.739 & 0.445 \\
%         \midrule
%         RHTex (Ours)
%         & \textbf{5000} & \textbf{4898}
%         & \textbf{40.748} & \textbf{1.302} & \textbf{98.595}
%         & \textbf{98.854} & \textbf{0.420} \\
%         \bottomrule
%     \end{tabular}
% \end{table*}

We then compute the components of the parent wave vector along the child's tangent basis:
\begin{equation}
    \mathbf{v}_{s,j}
    =
    \mathbf{B}_s^{\mathsf T}\mathbf{k}_{i,j},
    \qquad
    \mathbf{v}_{s,j}\in\mathbb{R}^2.
    \label{eq:child_wave_projection}
\end{equation}
To match the parent's phase variation along the child's tangent plane, we require $\frac{2\pi f_{s,j}}{\ell_s}\mathbf{d}_{s,j}=\mathbf{v}_{s,j}$. For $\|\mathbf{v}_{s,j}\|_2>0$, separating the magnitude and unit direction gives
\begin{equation}
\begin{aligned}
    f_{s,j}
    &= \frac{\ell_s}{2\pi}\|\mathbf{v}_{s,j}\|_2,\\
    \mathbf{d}_{s,j}
    &= \frac{\mathbf{v}_{s,j}}{\|\mathbf{v}_{s,j}\|_2}.
\end{aligned}
\label{eq:child_frequency_direction}
\end{equation}
In our implementation, projected frequencies are clamped to a small positive lower bound for numerical stability, and the inherited direction is retained when $\|\mathbf{v}_{s,j}\|_2=0$. When this frequency bound is inactive, the resulting child wave vector is $\mathbf{k}_{s,j}=\mathbf{B}_s\mathbf{B}_s^{\mathsf T}\mathbf{k}_{i,j}$, the orthogonal projection of the parent wave vector onto the child's tangent plane.

To account for the shift in coordinate origin, we set the child's phase offset to the parent's phase evaluated at the child center:
\begin{equation}
\begin{aligned}
    \beta_{s,j}
    &= \psi_{i,j}(\mathbf{p}_s)\\
    &= \beta_{i,j}
    + \mathbf{k}_{i,j}^{\mathsf T}
      (\mathbf{p}_s-\mathbf{p}_i),
\end{aligned}
\label{eq:child_phase_transfer}
\end{equation}
where phase offsets are understood modulo $2\pi$. The remaining appearance parameters are inherited from the parent: $\mathbf{c}_s=\mathbf{c}_i$, $\mathbf{a}_{s,j}=\mathbf{a}_{i,j}$, $\mathbf{b}_{s,j}=\mathbf{b}_{i,j}$, and $\rho_{s,j}=\rho_{i,j}$. Since both rational components are $2\pi$-periodic in their phase argument, this initialization preserves the parent's local appearance at the child center: $\mathbf{c}_s(\mathbf{p}_s)=\mathbf{c}_i(\mathbf{p}_s)$.

Together, wave-vector projection and phase correction align the child's inherited appearance with the parent's field under the local tangent-plane approximation. \cref{fig:split} compares this initialization with direct parameter copying, illustrating closer agreement with the parent-field reference. All harmonic orders are transferred jointly through their dependence on the same underlying phase function, without harmonic-specific adjustments. Wave directions are assigned analytically during child initialization and remain non-trainable, while frequencies and phase offsets continue to be optimized.

\subsection{Optimization}
\label{sec:optimization}

Following HKTex~\cite{foti2026heat}, we fit the representation to a reference UV texture by minimizing the photometric error at each sampled surface point $\mathbf{x}$:
\begin{equation}
    \mathcal{L}_{\mathrm{UV}}(\mathbf{x})
    =
    \left\|
        \mathbf{C}(\mathbf{x})
        -
        \mathbf{C}_{\mathrm{ref}}(\mathbf{x})
    \right\|_2^2,
    \label{eq:uv_loss}
\end{equation}
where $\mathbf{C}_{\mathrm{ref}}(\mathbf{x})$ is sampled from the reference texture at the corresponding UV coordinates.

For multi-view supervision, we optimize the surface color field through differentiable rendering with known camera poses and fixed mesh geometry and illumination. The photometric loss for a sampled camera ray $r$ is
\begin{equation}
    \mathcal{L}_{\mathrm{MV}}(r)
    =
    \left\|
        \widehat{\mathbf{I}}(r)
        -
        \mathbf{I}_{\mathrm{ref}}(r)
    \right\|_1,
    \label{eq:multiview_loss}
\end{equation}
where $\widehat{\mathbf{I}}(r)$ and $\mathbf{I}_{\mathrm{ref}}(r)$ denote the rendered and reference RGB values after identical clipping and photometric transformations. In each supervision setting, we minimize the corresponding loss averaged over a mini-batch of surface points or camera rays.

\section{Experiments}
\label{sec:experiments}

\begin{figure*}[t]
    \centering
    \includegraphics[width=\linewidth]{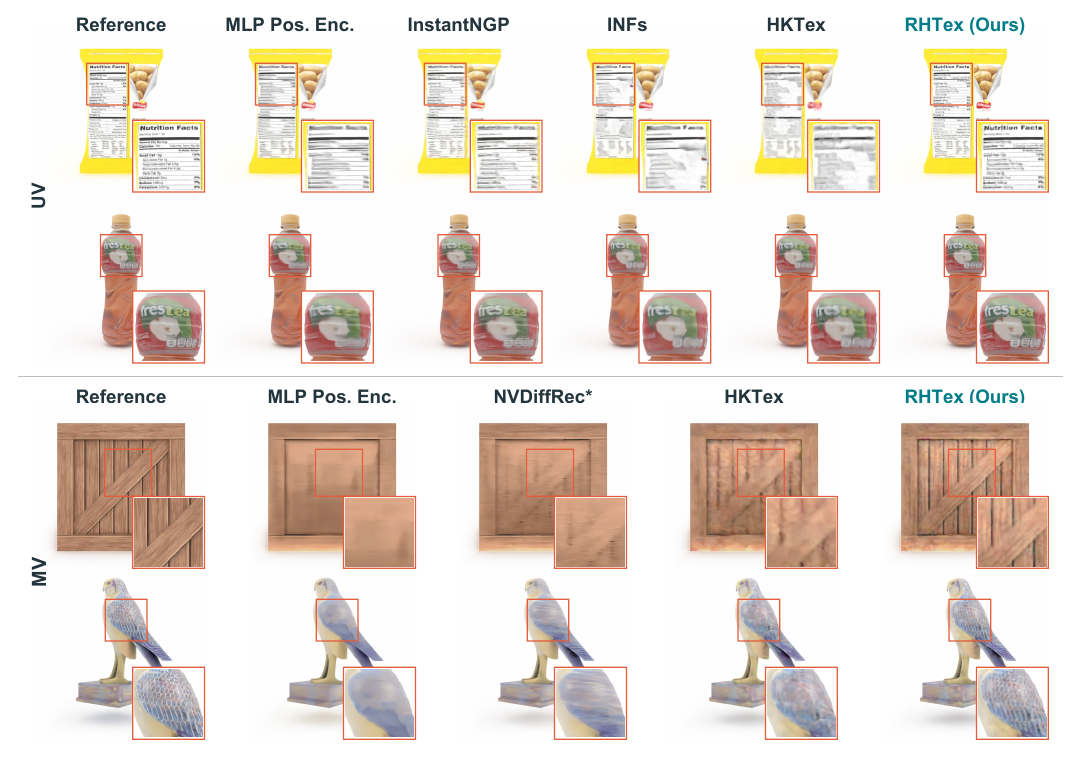}
    \caption{Qualitative comparison under UV-texture fitting and multi-view supervision. Enlarged regions show that our method better preserves fine text and high-frequency surface patterns. Additional qualitative results are provided in the supplementary material.}
    \label{fig:qualitative}
\end{figure*}

\paragraph{Implementation details.}
We follow HKTex's~\cite{foti2026heat} task-specific evaluation settings. UV-texture and multi-view fitting use 1,000 and 5,000 initial kernels, respectively. Our appearance model uses 39 learnable scalars per kernel, together with four non-trainable direction states. Frequencies are initialized to $0.5$, while harmonic RGB coefficients, phase offsets, and raw shape parameters are initialized to zero. Frequencies use a softplus parameterization with an additive floor of $10^{-4}$, and harmonic RGB coefficients are projected to $[-1,1]$ after each update.
% The added appearance parameters are optimized using Adam with initial learning rates of $6.1543\times10^{-3}$ for UV fitting and $10^{-3}$ for multi-view fitting, following the respective learning-rate schedules.

\paragraph{Evaluation setup and metrics.}
We use a filtered subset of Objaverse~\cite{deitke2023objaverse}, following the data filtering procedure of HKTex~\cite{foti2026heat}. We independently sample 300 meshes at random for UV-texture fitting and multi-view fitting, respectively. All methods optimize only surface albedo on fixed meshes, with known camera poses and fixed illumination for multi-view fitting. We use HKTex as the primary baseline and its neural-texture implementations for additional comparisons: MLP Pos. Enc. for both tasks, InstantNGP~\cite{muller2022instant} and INFs~\cite{koestler2022intrinsic} for UV-texture fitting, and NVDiffRec\textsuperscript{*}~\cite{munkberg2022extracting} for multi-view fitting. The superscript $*$ denotes HKTex's albedo-only adaptation of NVDiffRec. We compare reconstructed and reference renderings from five viewpoints at $256\times256$ resolution, reporting PSNR, LPIPS, SSIM, MS-SSIM, and MSE averaged across objects. Time reports mean optimization wall time. Storage reports mean compressed appearance-state size (KB).

\subsection{Comparative studies}
\label{sec:comparative}

\begin{figure}[t]
    \centering
    \includegraphics[width=\linewidth]{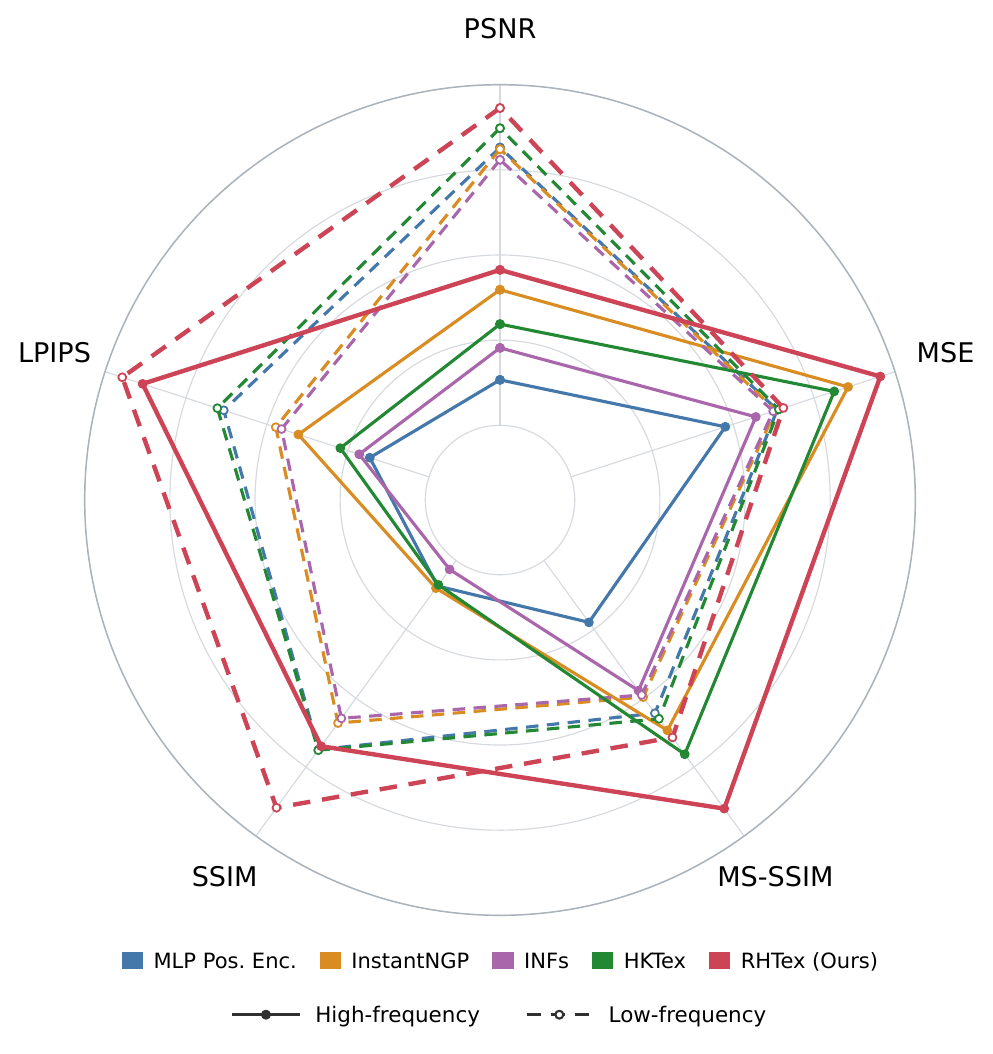}
    \caption{UV-texture fitting on selected high- and low-frequency texture groups.}
    \label{fig:uv_frequency_comparison}
\end{figure}

\paragraph{UV-texture fitting.}
RHTex achieves the best results across all five reconstruction metrics (\cref{tab:uv_multiview_results}). Compared with HKTex, it increases mean PSNR from $40.297$ to $40.748$~dB and reduces LPIPS by $30.1\%$. These gains are accompanied by a modest increase in recorded optimization time from $2.57$ to $2.87$ minutes and higher storage. \cref{fig:qualitative} further shows improved reconstruction of fine, high-frequency texture details.

To test whether a larger budget alleviates HKTex's oversmoothing, we vary its kernel cap from 5,000 to 10,000 in increments of 1,000. At a cap of 10,000, HKTex's mean final kernel count rises from 4,834 to 8,578, while PSNR improves only from $40.297$ to $40.349$~dB. RHTex achieves $40.748$~dB with 4,898 kernels on average under a cap of 5,000 (\cref{tab:uv_kernel_limit}). These results suggest that enriching per-kernel appearance is more effective than increasing the number of constant-color kernels under the evaluated optimization settings.

We further compare methods on two selected groups of 80 meshes each, with high- and low-frequency textures, for UV fitting (\cref{fig:uv_frequency_comparison}). Relative to HKTex, RHTex improves PSNR by $0.956$ and $0.356$~dB and reduces LPIPS by $43.82\%$ and $28.97\%$ in the high- and low-frequency groups, respectively, indicating larger gains on the high-frequency cases.

\paragraph{Multi-view fitting.}
We fit surface albedo from rendered observations with known camera poses and fixed geometry and illumination. RHTex improves all five reconstruction metrics over HKTex (\cref{tab:uv_multiview_results}), increasing mean PSNR from $35.232$ to $35.646$~dB and reducing LPIPS by $24.6\%$. The qualitative results (\cref{fig:qualitative}) also show improved reconstruction quality, consistent with the UV results.

\subsection{Ablation study}
\label{sec:ablation}

\begin{table}[t]
    \centering
    \caption{Ablation of rational harmonics and frame-aware harmonic reparameterization for UV-texture fitting. Bold indicates the best result.}
    \label{tab:rho_inheritance_ablation}
    \resizebox{\columnwidth}{!}{%
    \begin{tabular}{lccccc}
        \toprule
        Variant
        & \shortstack{PSNR \\ $(\uparrow)$}
        & \shortstack{LPIPS \\ $(\times 10^{-2}, \downarrow)$}
        & \shortstack{SSIM \\ $(\times 10^{-2}, \uparrow)$}
        & \shortstack{MS-SSIM \\ $(\times 10^{-2}, \uparrow)$}
        & \shortstack{MSE \\ $(\times 10^{-3}, \downarrow)$} \\
        \midrule
        Constant
        & 40.297 & 1.863 & 98.208 & 98.715 & 0.447 \\
        $\rho = 0$, w/o reparameterization
        & 40.564 & 1.401 & 98.527 & 98.825 & 0.429 \\
        $\rho = 0$, w/ reparameterization
        & 40.616 & 1.394 & 98.531 & 98.831 & 0.427 \\
        Learnable $\rho$, w/o reparameterization
        & 40.691 & 1.319 & 98.582 & 98.846 & 0.422 \\
        Full (Ours)
        & \textbf{40.748} & \textbf{1.302} & \textbf{98.595}
        & \textbf{98.854} & \textbf{0.420} \\
        \bottomrule
    \end{tabular}%
    }
\end{table}

\begin{figure}[t]
    \centering
    \includegraphics[width=\linewidth]{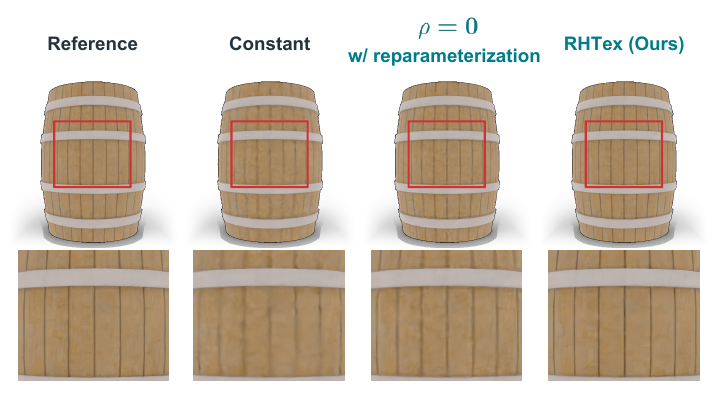}
    \caption{Qualitative comparison of local appearance models for UV-texture fitting. Both harmonic variants use frame-aware harmonic reparameterization.}
    \label{fig:ab_qualitative}
\end{figure}

We ablate the rational harmonic appearance field and frame-aware harmonic reparameterization. Our appearance representations form a nested hierarchy: setting $\rho=0$ retains only fundamental sinusoids, and further zeroing the sine and cosine RGB coefficients yields constant-color kernels. With HKTex as the constant-color reference, we compare direct parameter copying and reparameterization for both the $\rho=0$ and learnable-$\rho$ variants (\cref{tab:rho_inheritance_ablation}).

With reparameterization disabled, replacing constant per-kernel color with sinusoidal appearance ($\rho=0$) increases PSNR from $40.297$ to $40.564$~dB and reduces LPIPS from $1.863$ to $1.401$ ($\times10^{-2}$). Learning $\rho$ further improves these metrics to $40.691$~dB and $1.319$, respectively. These progressive gains highlight the value of replacing constant per-kernel colors with spatially varying local appearance fields. Rational harmonics further enrich these fields through higher-order components beyond fundamental sinusoids, translating increased per-kernel expressiveness into improved texture reconstruction. The qualitative comparison (\cref{fig:ab_qualitative}) is consistent with these quantitative results, showing finer texture details and reduced smearing with rational harmonics.

Frame-aware harmonic reparameterization reduces appearance misalignment caused by center shifts and tangent-frame changes during splitting. At child initialization, it better matches the parent-derived reference under identical child heat weights and blend normalization (\cref{fig:split}). Even with enhanced per-kernel expressiveness, the aggregate ablation results show further gains across all five metrics for both the $\rho=0$ and learnable-$\rho$ variants (\cref{tab:rho_inheritance_ablation}). These results highlight the complementary roles of expressive local appearance and consistent split initialization in improving reconstruction. Further analysis is provided in the supplementary material.

% Frame-aware harmonic reparameterization reduces appearance misalignment caused by center shifts and tangent-frame changes during splitting. At child initialization, it better matches the parent-derived reference under identical child heat weights and blend normalization (\cref{fig:split}). The aggregate ablation results favor reparameterization across all five metrics for both the $\rho=0$ and learnable-$\rho$ variants (\cref{tab:rho_inheritance_ablation}). Further analysis is provided in the supplementary material.

% Although reparameterization reduces split-induced appearance discrepancies, its average reconstruction gains are modest (\cref{tab:rho_inheritance_ablation}). Our rational harmonic model's enhanced per-kernel expressiveness may allow subsequent optimization to compensate for these discrepancies. To examine when reparameterization is most beneficial, we further select high- and low-frequency texture groups, each comprising 80 meshes. The high-frequency group shows larger gains in most metrics (\cref{fig:split_reparameterization}), suggesting that preserving learned appearance during splitting is particularly beneficial for recovering fine details.

\section{Conclusion}
\label{sec:conclusion}

We presented Rational Harmonic Textures for detailed surface appearance reconstruction. Our results suggest that enriching per-kernel appearance can be more effective than increasing kernel density alone. Maintaining local appearance consistency during splitting provides a complementary benefit, highlighting the value of jointly designing local appearance representations and their adaptive optimization.

% We presented Rational Harmonic Textures, which enrich surface heat kernels with compact local appearance fields represented by rational trigonometric atoms. Each atom encodes a harmonic series without independently parameterizing each harmonic order, enabling richer within-kernel color variation. We also derived an analytic reparameterization of wave vectors and phase offsets that accounts for center translations and tangent-frame changes when initializing split children. Experiments on the evaluated UV-texture fitting and multi-view subsets demonstrate improved reconstruction of fine surface details relative to HKTex.

% \section{Final copy}

% You must include your signed IEEE copyright release form when you submit your finished paper.
% We MUST have this form before your paper can be published in the proceedings.

% Please direct any questions to the production editor in charge of these proceedings at the IEEE Computer Society Press:
% \url{https://www.computer.org/about/contact}.
\clearpage

% \subsection*{LLM Usage Statement}
% We used OpenAI Codex to assist with drafting and revising passages in the method, experiments, and conclusion sections, as well as captions and \LaTeX{} formatting. It also assisted with code review, consistency checks between the implementation and manuscript, experimental result analysis, and figure and table presentation. The authors reviewed the AI-assisted material and take full responsibility for the final manuscript, implementation, experimental results, and scientific claims.

{
    \small
    \bibliographystyle{ieeenat_fullname}
    \bibliography{main}
}


\end{document}